%% file: ifacconf.tex
\documentclass{ifacconf}
\usepackage{siunitx}
\usepackage{graphicx} 
\usepackage{transparent}
\usepackage{natbib}  
\graphicspath{{./}{images/}}
\usepackage{import}
\usepackage{lipsum}
\usepackage{amsfonts}
\usepackage{bm}
\usepackage{xspace} %
\usepackage{amsmath}
\usepackage{xcolor}
\usepackage{epstopdf}
\usepackage{subcaption}  

\newcommand{\FullStop}{\nobreakspace.\xspace}

\newcommand{\Comma}{\nobreakspace,\xspace}

\begin{document}
\begin{frontmatter}

\title{Learning Swing-up Maneuvers for a Suspended Aerial Manipulation Platform in a Hierarchical Control Framework} 
\author[First]{Hemjyoti Das} 
\author[First,Second]{Minh Nhat Vu} 
\author[First,Third]{Christian Ott}

\address[First]{Automation and Control Institute (ACIN), TU Wien, Gusshausstraße
27-29, 1040, Vienna, Austria (email: hemjyoti.das@tuwien.ac.at,  minh.vu@tuwien.ac.at, christian.ott@tuwien.ac.at)}
\address[Second]{Austrian Institute of Technology (AIT) GmbH, 1210, Vienna, Austria}
\address[Third]{Institute of Robotics and Mechatronics, German Aerospace Center
(DLR), Oberpfaffenhofen, Muenchener Strasse 20, 82234, Wessling, Germany }

\begin{abstract}           
\import{sections/}{0-abstract}
\end{abstract}

\begin{keyword}
Aerial manipulation, swing-up control, hierarchical control, reinforcement learning
\end{keyword}
\end{frontmatter}

\section{Introduction}
\import{sections/}{1-introduction}

\section{System Dynamics and Overview}
\import{sections/}{2-dynamics}

\section{Controller Design}
\import{sections/}{3-controller_design}

\section{Learning Policy and Task Reference}
\import{sections/}{4-rl}

\section{Results}
\import{sections/}{5-experiments}

\section{Conclusion}
\import{sections/}{6-conclusion}

\bibliography{ifacconf}       
\end{document}

%% file: sections/0-abstract.tex
In this work, we present a novel approach to augment a model-based control method with a reinforcement learning (RL) agent and demonstrate a swing-up maneuver with a suspended aerial manipulation platform. These platforms are targeted towards a wide range of applications on construction sites involving cranes, with swing-up maneuvers allowing it to perch at a given location, inaccessible with purely the thrust force of the platform. Our proposed approach is based on a hierarchical control framework, which allows different tasks to be executed according to their assigned priorities. An RL agent is then subsequently utilized to adjust the reference set-point of the lower-priority tasks to perform the swing-up maneuver, which is confined in the nullspace of the higher-priority tasks, such as maintaining a specific orientation and position of the end-effector. Our approach is validated using extensive numerical simulation studies. 

%% file: sections/1-introduction.tex
\label{sec:intro}
Aerial robotic manipulation has advanced significantly in recent years for complex physical-interaction tasks such as inspection and maintenance of pipelines, electricity lines, and remotely located construction sites (\cite{ruggiero2018aerial, ollero2021past,  ryll20176d}). These systems, usually comprising of a multirotor platform equipped with a manipulator (\cite{ollero2021past}), typically demand a high amount of energy for the stabilization of the complete system, which further leads to a reduced flight time and payload capacity (\cite{paredes2017study,villa2020survey}). Additionally, the large size of the propellers necessary to compensate for the mass of the heavy manipulator can cause turbulence in the surroundings (\cite{kondak2014aerial}). \par 
To tackle these challenges, a suspended aerial manipulator (SAM) platform is proposed in \cite{sarkisov2019development}, which consists of an omnidirectional platform suspended from a carrier system using rigid cables. The SAM platform is equipped with a 7 Degrees of Freedom (DoF) manipulator. It has been utilized to demonstrate compliant physical interaction with an unknown environment in a hierarchical control framework (\cite{coelho2021hierarchical, sarkisov2023hierarchical, gabellieri2020compliance}). In \cite{yiugit2021novel, kong2022robotic}, similar suspended aerial manipulator platforms have been proposed to demonstrate accurate end-effector trajectory tracking and teleoperation tasks with the environment. \par
 \begin{figure}[t
 ]
    \centering
\includegraphics[width = 0.3\textwidth]{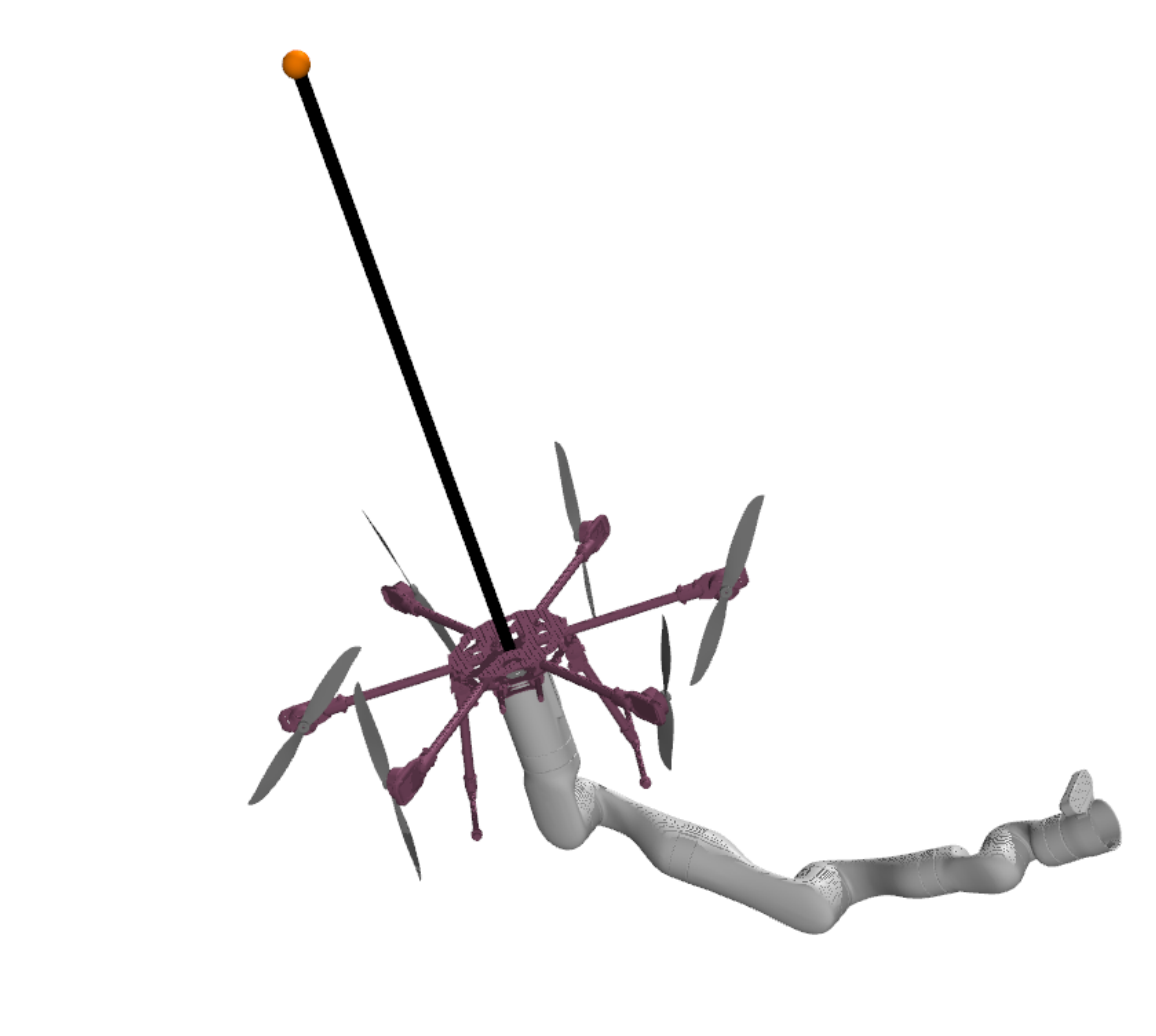}
    \caption{Suspended aerial manipulation platform.} \label{fig:plat_kinnova}
\end{figure}


This work extends the application of such a suspended aerial platform for performing swing-up maneuvers by utilizing the motion of only the manipulator. The problem of generating swing-up motion for different mechanical systems has been widely studied in the literature. For instance, during the 1960s, research into stabilizing single inverted pendulums gained significant attention, as demonstrated in works like  \cite{cannon1967dynamics} and \cite{ogata1970modern}. These studies primarily relied on specially designed mechanical systems that generate the necessary forces and torques for swing-up and stabilization tasks. However, implementing such tasks on industrial robots presents additional challenges, including constraints on joint positions, velocities, and torques. \cite{winkler2009erecting} implemented a combined feedforward and balancing control strategy to achieve the swing-up and stabilization of a single inverted pendulum on a KUKA KR3 robot. More recently, advancements in the control of robotic arms with seven degrees of freedom (DoFs) have utilized Bayesian optimization ( \cite{marco2016automatic}) and model-based policy search (\cite{doerr2017model}) to automate the tuning of control strategies for balancing tasks. The stabilization of more complex pendulums, such as spherical pendulums, is first successfully demonstrated on the ``DLR LWR II" robot by \cite{schreiber2001interactive}. Building on this foundation, recent studies have begun exploring the swing-up of spherical pendulums using industrial robots, with promising experimental results reported in \cite{vu2021fast}.  \par 
This work considers such swing-up maneuvers for a suspended aerial manipulation platform. However, our work aims to accomplish other tasks with the platform while simultaneously performing the swing-up maneuver, for which we utilize a hierarchical control framework (\cite{dietrich2019hierarchical}). Thus, one of the contributions of this work is to formulate a task-hierarchy approach, where we ensure that the end-effector always maintains a specific orientation and position along a particular axis with a given yaw orientation of the platform. This is motivated by the choice of the mounted camera system on both the platform and end-effector of the manipulator, which should always observe the given target to grasp once the swing-up motion is completed. Concurrently, the swing-up control is performed as a second-priority task in a dynamically decoupled manner. A reinforcement learning agent is then proposed to adjust the reference coordinates for the lower-priority tasks of the designed hierarchical whole-body control framework and thus successfully learn the swing-up maneuver in the nullspace of the higher-priority tasks, which is the main contribution of this paper. The effectiveness of our proposed approaches is validated using extensive simulation studies. \par 
The rest of the paper is organized as follows. In Section \ref{sec:sys}, we present the system dynamics of the suspended aerial platform with the attached manipulator. Section \ref{sec:control} introduces the task definition and hierarchical control formulation. The reinforcement learning policy and task references necessary for the swing-up maneuver are presented in Section \ref{sec:rl}. Finally, the simulation results are shown in Section \ref{sec:exp}, while Section \ref{sec:con} concludes the paper.

%% file: sections/2-dynamics.tex
\label{sec:sys}
\begin{figure}[t]
    \centering
\def\svgwidth{0.8\columnwidth}
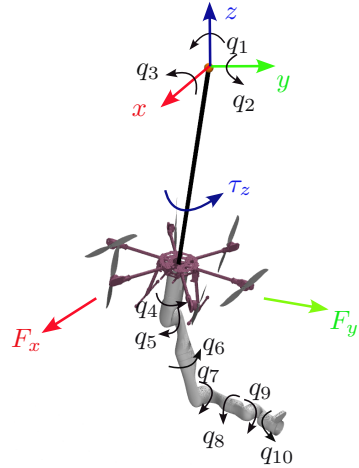
\vspace{-20pt}
    \caption{Schematic representation of the cable-suspended aerial manipulation platform.}
\label{fig:platform_schematics}
\end{figure}
In this work, we performed the swing-up maneuvers using the planar-thrust suspended aerial platform previously developed by our group in \cite{das2023}. The oscillations of the suspended platform are modeled as a single spherical pendulum, similar to the work by  \cite{sarkisov2020optimal}, with the cable used for the suspension assumed to be rigid and taut. As seen from Fig. \ref{fig:platform_schematics}, we denote the generalized coordinates of the suspended platform as $q_1$, $q_2$ and $q_3$ which are the rotations of the suspension cable about its $z$, $y$ and $x$ axis, respectively. The platform can apply a planar wrench $\mathbf{u}_p$ about its axis (\cite{das2023}), which consists of the translational forces along its $x$ and $y$ axis, denoted as $F_x$ and $F_y$, respectively, and the torque $\tau_z$ exerted about its $z$ axis. The planar wrench is related to the exerted joint torque $\bm{\tau}_p$ at the platform joints using Jacobian $\mathbf{J}_u \in \mathbb{R}^{3 \times 3}$ as  
\begin{equation}
    \bm{\tau}_p = \mathbf{J}_u^T \mathbf{u}_p
\end{equation}\par 
In this work, we considered the 7-DoF Kinnova Gen3 robotic manipulator attached to the suspended platform. Our framework is highly versatile and not restricted to a specific manipulator, allowing compatibility with a wide range of manipulators attached to the suspended platform. The generalized coordinates of the manipulator is denoted as $\mathbf{q}_m \in \mathbb{R}^7$, which consists of the joints $q_4$, $q_5$, $q_6$, $q_7$, $q_8$, $q_9$ and $q_{10}$ as shown in Fig. \ref{fig:platform_schematics}. The joint torque exerted by the manipulator is denoted by the vector $\bm{\tau}_m \in \mathbb{R}^7$. The complete suspended aerial manipulation system is thus modeled as a 10 DoF system, with its system dynamics summarized as 
\begin{equation}
\label{eq:dyn}
\mathbf{M\left(q\right)}\mathbf{\ddot{q}} +  \mathbf{C\left(q, \dot{q}\right)}\mathbf{\dot{q}} +  \mathbf{g\left(q\right)} =  \bm{\tau} \Comma
\end{equation} 
where $\mathbf{M\left(q\right)} \in \mathbb{R}^{10 \times 10}$ is the inertia matrix, $\mathbf{C\left(q, \dot{q}\right)} \in \mathbb{R}^{10 \times 10}$, $\mathbf{g\left(q\right)} \in \mathbb{R}^{10}$ is the gravity torque exerted on each joint. $\bm{\tau}$ is the generalized joint torque vector summarized as 
\begin{equation}
\label{eq:tau}
\bm{\tau} = \begin{bmatrix}
        \bm{\tau}_p & \bm{\tau}_m 
    \end{bmatrix}^{T} 
\end{equation} \par 

%% file: images/plat_schematics_2.pdf_tex
\begingroup%
  \makeatletter%
  \providecommand\color[2][]{%
    \errmessage{(Inkscape) Color is used for the text in Inkscape, but the package 'color.sty' is not loaded}%
    \renewcommand\color[2][]{}%
  }%
  \providecommand\transparent[1]{%
    \errmessage{(Inkscape) Transparency is used (non-zero) for the text in Inkscape, but the package 'transparent.sty' is not loaded}%
    \renewcommand\transparent[1]{}%
  }%
  \providecommand\rotatebox[2]{#2}%
  \newcommand*\fsize{\dimexpr\f@size pt\relax}%
  \newcommand*\lineheight[1]{\fontsize{\fsize}{#1\fsize}\selectfont}%
  \ifx\svgwidth\undefined%
    \setlength{\unitlength}{780.42644831bp}%
    \ifx\svgscale\undefined%
      \relax%
    \else%
      \setlength{\unitlength}{\unitlength * \real{\svgscale}}%
    \fi%
  \else%
    \setlength{\unitlength}{\svgwidth}%
  \fi%
  \global\let\svgwidth\undefined%
  \global\let\svgscale\undefined%
  \makeatother%
  \begin{picture}(1,1.04245005)%
    \lineheight{1}%
    \setlength\tabcolsep{0pt}%
    \put(0,0){\includegraphics[width=\unitlength,page=1]{plat_schematics_2.pdf}}%
    \put(0.65160494,0.38239494){\color[rgb]{0,0.91764706,0.03529412}\transparent{0.94851202}\makebox(0,0)[lt]{\lineheight{1.25}\smash{\begin{tabular}[t]{l}$F_y$\end{tabular}}}}%
    \put(0.55292379,0.83666831){\color[rgb]{0,0.91764706,0.03529412}\transparent{0.94851202}\makebox(0,0)[lt]{\lineheight{1.25}\smash{\begin{tabular}[t]{l}$y$\end{tabular}}}}%
    \put(0.29249417,0.85804854){\color[rgb]{0,0,0.00392157}\transparent{0.94851202}\makebox(0,0)[lt]{\lineheight{1.25}\smash{\begin{tabular}[t]{l}$q_3$\end{tabular}}}}%
    \put(0.47142298,0.79841308){\color[rgb]{0.00392157,0,0.01568627}\transparent{0.94851202}\makebox(0,0)[lt]{\lineheight{1.25}\smash{\begin{tabular}[t]{l}$q_2$\end{tabular}}}}%
    \put(0.46307805,0.64055592){\color[rgb]{0,0.03137255,0.91764706}\transparent{0.94851202}\makebox(0,0)[lt]{\lineheight{1.25}\smash{\begin{tabular}[t]{l}$\tau_z$\end{tabular}}}}%
    \put(0.45681474,0.96339862){\color[rgb]{0,0.03137255,0.91764706}\transparent{0.94851202}\makebox(0,0)[lt]{\lineheight{1.25}\smash{\begin{tabular}[t]{l}$z$\end{tabular}}}}%
    \put(0.06025478,0.34980451){\color[rgb]{0.91764706,0,0.04705882}\transparent{0.94851202}\makebox(0,0)[lt]{\lineheight{1.25}\smash{\begin{tabular}[t]{l}$F_x$\end{tabular}}}}%
    \put(0.28418764,0.78267996){\color[rgb]{0.91764706,0,0.04705882}\transparent{0.94851202}\makebox(0,0)[lt]{\lineheight{1.25}\smash{\begin{tabular}[t]{l}$x$\end{tabular}}}}%
    \put(0,0){\includegraphics[width=\unitlength,page=2]{plat_schematics_2.pdf}}%
    \put(0.45985241,0.90767779){\color[rgb]{0.01176471,0.00784314,0.01960784}\transparent{0.94851202}\makebox(0,0)[lt]{\lineheight{1.25}\smash{\begin{tabular}[t]{l}$q_1$\end{tabular}}}}%
    \put(0.29066994,0.41307361){\color[rgb]{0.01568627,0.01176471,0.02745098}\transparent{0.94851202}\makebox(0,0)[lt]{\lineheight{1.25}\smash{\begin{tabular}[t]{l}$q_4$\end{tabular}}}}%
    \put(0.28924117,0.35433251){\color[rgb]{0.01176471,0.00784314,0.02352941}\transparent{0.94851202}\makebox(0,0)[lt]{\lineheight{1.25}\smash{\begin{tabular}[t]{l}$q_5$\end{tabular}}}}%
    \put(0.41583791,0.34281132){\color[rgb]{0.00784314,0.00392157,0.01568627}\transparent{0.94851202}\makebox(0,0)[lt]{\lineheight{1.25}\smash{\begin{tabular}[t]{l}$q_6$\end{tabular}}}}%
    \put(0.40550275,0.29062879){\color[rgb]{0.00784314,0.00392157,0.01960784}\transparent{0.94851202}\makebox(0,0)[lt]{\lineheight{1.25}\smash{\begin{tabular}[t]{l}$q_7$\end{tabular}}}}%
    \put(0.41538718,0.16968489){\color[rgb]{0.00784314,0,0.01960784}\transparent{0.94851202}\makebox(0,0)[lt]{\lineheight{1.25}\smash{\begin{tabular}[t]{l}$q_8$\end{tabular}}}}%
    \put(0.50150751,0.25989908){\color[rgb]{0.01176471,0.00784314,0.02352941}\transparent{0.94851202}\makebox(0,0)[lt]{\lineheight{1.25}\smash{\begin{tabular}[t]{l}$q_9$\end{tabular}}}}%
    \put(0.52252191,0.13881221){\color[rgb]{0.00392157,0,0.01960784}\transparent{0.94851202}\makebox(0,0)[lt]{\lineheight{1.25}\smash{\begin{tabular}[t]{l}$q_{10}$\end{tabular}}}}%
  \end{picture}%
\endgroup%

%% file: sections/3-controller_design.tex
\label{sec:control}
\begin{figure*}[t]
    \centering
\def\svgwidth{1.75\columnwidth}
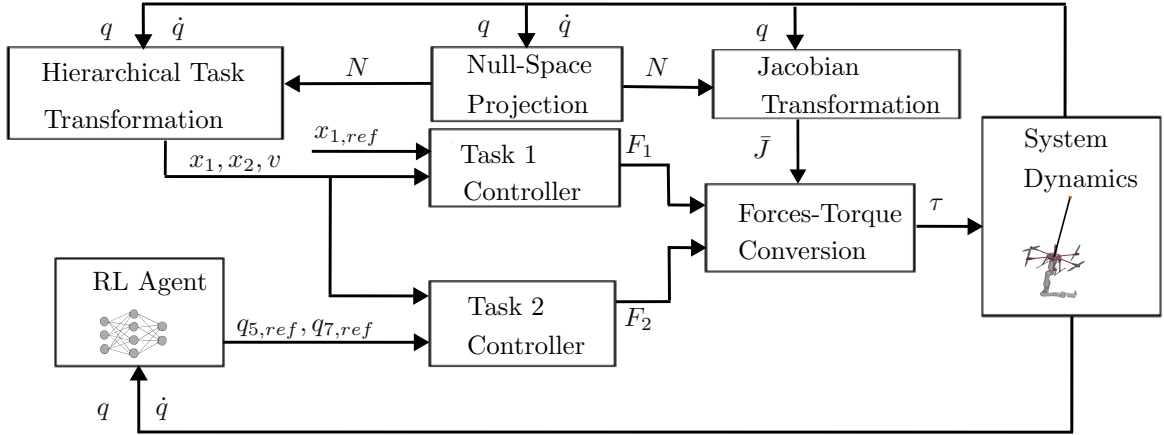
    \caption{Overview of our proposed control framework.}
\label{fig:controller_block}
\end{figure*}
In this section, we design a whole-body controller in a hierarchical framework. Such control approaches (\cite{dietrich2019hierarchical}) allow us to execute multiple tasks while assigning a certain priority to each task, such that the lower-priority tasks do not interfere with the execution of the higher-priority tasks. This strict hierarchy between different tasks is achieved by defining dynamically consistent null-space projectors, which leads to decoupled task-space velocities. In this work, we define the highest priority task $\mathbf{x}_1 \in \mathbb{R}^5$ as the stabilization of the end-effector orientation, end-effector position along the axis perpendicular to the swing-up axis, and the platform yaw angle. The second priority task $\mathbf{x}_2\in \mathbb{R}^5$ concerns the motion of the specific joints, which enable the swing-up of the complete system. $\mathbf{x}_2$ also contains the platform state $q_2$ along the swing-up axis, as will be explained in the subsequent section. The task-spaces $\mathbf{x}_1$ and $\mathbf{x}_2$ are thus summarized as follows, 
\begin{equation}
\label{eq:task}
\begin{aligned}
    \mathbf{x}_1 &=  \begin{bmatrix}
       y_e & \phi & \theta & \psi & q_1
    \end{bmatrix}^{T}\\
    \mathbf{x}_2 &= \begin{bmatrix}
    q_2 &  q_4 & q_5 & q_6 & q_7 
    \end{bmatrix}^{T}   \Comma
\end{aligned} 
\end{equation}
where $y_e$ is the end-effector's position along its $y$ axis. $\phi$, $\theta$ and $\psi$ are the Euler angles of the end-effector defined about its $x$, $y$ and $z$ axis, respectively. Note that the goal of this work is to perform the swing-up maneuver along the $x$-$z$ plane, as shown in Fig. \ref{fig:platform_schematics} while keeping the end-effector along a given orientation and a given $y$-axis position, in order to observe the given target to be grasped after swinging up. Additionally, the first-priority task includes the yaw orientation $q_1$ for similar reasons and thus should always point in a specific direction to observe the target. The second priority task $\mathbf{x}_2$ concerns the motion of certain joints, which will aid in the swing-up maneuver. Note that in this work, we aim to swing up the complete system by primarily utilizing the motion of the manipulator while using minimal propulsion force from the platforms, to ensure energy efficiency. Based on these definitions, we summarize the task-space velocities as 
\begin{equation}
\label{eq:task_vel}
\begin{aligned}
    \mathbf{\dot{x}}_1 &= \mathbf{J}_1 \mathbf{\dot{q}} \\ 
    \mathbf{\dot{x}}_2 &= \mathbf{J}_2 \mathbf{\dot{q}} \Comma 
\end{aligned}
\end{equation}
where the the task-space Jacobians $\mathbf{J}_1 \in \mathbb{R}^{5 \times 10}$ and $\mathbf{J}_2 \in \mathbb{R}^{5 \times 10}$ are obtained kinematically and have full row-rank. Next, to decouple the task-space velocities and to define a strict hierarchy amongst the different tasks, we define the null-space projection matrix (\cite{dietrich2019hierarchical}) for the second task as
\begin{equation}
\label{eq:null}
\begin{aligned}
    \mathbf{N} = \mathbf{I} - \mathbf{J}_1^T\mathbf{J}_1^{M+,T}  \Comma 
\end{aligned}
\end{equation}
where $\mathbf{I}\in \mathbb{R}^{10 \times 10}$ is the identity matrix, while the operator $^{M+}$ denotes a dynamically consistent pseudoinversion \cite{dietrich2019hierarchical}. Next, we define the decoupled Jacobian matrices $\mathbf{\bar{J}}$ as 
\begin{equation}
\label{eq:decouple_J}
\mathbf{\bar{J}} = \begin{bmatrix}
    \mathbf{\bar{J}}_1 \\ 
    \mathbf{\bar{J}}_2 
\end{bmatrix} =  \begin{bmatrix}
    \mathbf{J}_1 \\ 
    \mathbf{J}_2 \mathbf{N}^T  
\end{bmatrix} \FullStop
\end{equation} \par 
The decoupled task-space velocities are then defined as 
\begin{equation}
\label{eq:task_v}
\mathbf{v} = \begin{bmatrix}
      \mathbf{v}_1 \\ 
    \mathbf{v}_2 
\end{bmatrix} = \begin{bmatrix}
    \mathbf{\bar{J}}_1 \\ 
    \mathbf{\bar{J}}_2 
\end{bmatrix} \mathbf{\dot{q}} = \mathbf{\bar{J}}\mathbf{\dot{q}} \FullStop
\end{equation} \par 
By utilizing these task-space coordinates, we conduct the coordinate transformation of \eqref{eq:dyn} to obtain the hierarchical decoupled equations of motion as 
\begin{equation}
\label{eq:task_equation}
\bm{\Lambda}\left(\mathbf{q} \right) \mathbf{\dot{v}} + \bm{\mu}\left(\mathbf{q}, \mathbf{\dot{q}} \right) \mathbf{v} = \mathbf{\bar{J}}^{-T} \left( \bm{\tau} - \mathbf{g}  \right) \Comma
\end{equation}
where $\bm{\Lambda}$ = $\left(\mathbf{\bar{J}}\mathbf{M}^{-1}\mathbf{\bar{J}}^{T}\right)^{-1}$ is the decoupled task-space inertia and $\bm{\mu}$ = $\bm{\Lambda}\left( \mathbf{\bar{J}}\mathbf{M}^{-1}\mathbf{C}\mathbf{\bar{J}}^{-1} - \mathbf{\dot{\bar{J}}} \mathbf{\bar{J}}^{-1} \right)$ is the transformed Coriolis term in the task space. Next, we define the joint torque $\bm{\tau{}}$ as 
\begin{equation}
\label{eq:joint_torque}
\bm{\tau} = \mathbf{g} + \bm{\tau}_u + \mathbf{\bar{J}}^{T}\mathbf{F} \Comma   
\end{equation}
where the torque $\bm{\tau}_u$ compensates for the cross-coupling Coriolis components (\cite{dietrich2019hierarchical}), whereas the task-space force $\mathbf{F} = \begin{bmatrix}
    \mathbf{F}_1 & \mathbf{F}_{2}
\end{bmatrix}^T$ consists of the forces necessary to accomplish task 1 and task 2, defined by $\mathbf{F}_1$ and $\mathbf{F}_2$, respectively, and both are chosen as PD control laws, similar to \cite{das2023hardware}. The control law \eqref{eq:joint_torque} thus ensures the execution of the different tasks hierarchically. The control law also ensures that the swing-up maneuver will be executed primarily due to the motion of the manipulator by choosing an appropriate task force $\mathbf{F}_2$, as will be discussed subsequently.

%% file: images/controller_block_diagram.pdf_tex
\begingroup%
  \makeatletter%
  \providecommand\color[2][]{%
    \errmessage{(Inkscape) Color is used for the text in Inkscape, but the package 'color.sty' is not loaded}%
    \renewcommand\color[2][]{}%
  }%
  \providecommand\transparent[1]{%
    \errmessage{(Inkscape) Transparency is used (non-zero) for the text in Inkscape, but the package 'transparent.sty' is not loaded}%
    \renewcommand\transparent[1]{}%
  }%
  \providecommand\rotatebox[2]{#2}%
  \newcommand*\fsize{\dimexpr\f@size pt\relax}%
  \newcommand*\lineheight[1]{\fontsize{\fsize}{#1\fsize}\selectfont}%
  \ifx\svgwidth\undefined%
    \setlength{\unitlength}{3393.88160525bp}%
    \ifx\svgscale\undefined%
      \relax%
    \else%
      \setlength{\unitlength}{\unitlength * \real{\svgscale}}%
    \fi%
  \else%
    \setlength{\unitlength}{\svgwidth}%
  \fi%
  \global\let\svgwidth\undefined%
  \global\let\svgscale\undefined%
  \makeatother%
  \begin{picture}(1,0.40327923)%
    \lineheight{1}%
    \setlength\tabcolsep{0pt}%
    \put(0.87475027,0.25832129){\color[rgb]{0,0,0}\makebox(0,0)[lt]{\lineheight{1.25}\smash{\begin{tabular}[t]{l}  System \\\end{tabular}}}}%
    \put(0.87412187,0.25562978){\color[rgb]{0,0,0}\makebox(0,0)[lt]{\lineheight{1.25}\smash{\begin{tabular}[t]{l}\\Dynamics\end{tabular}}}}%
    \put(0.63177832,0.19405266){\color[rgb]{0,0,0}\makebox(0,0)[lt]{\lineheight{1.25}\smash{\begin{tabular}[t]{l}  Forces-Torque\\    Conversion\end{tabular}}}}%
    \put(0.40124395,0.32149293){\color[rgb]{0,0,0}\makebox(0,0)[lt]{\lineheight{1.25}\smash{\begin{tabular}[t]{l}  Null-Space \\\end{tabular}}}}%
    \put(0.40163105,0.31846762){\color[rgb]{0,0,0}\makebox(0,0)[lt]{\lineheight{1.25}\smash{\begin{tabular}[t]{l}\\  Projection\end{tabular}}}}%
    \put(0.03995285,0.31487924){\color[rgb]{0,0,0}\makebox(0,0)[lt]{\lineheight{1.25}\smash{\begin{tabular}[t]{l}  Hierarchical Task \\\end{tabular}}}}%
    \put(0.04190849,0.30569911){\color[rgb]{0,0,0}\makebox(0,0)[lt]{\lineheight{1.25}\smash{\begin{tabular}[t]{l}\\   Transformation\end{tabular}}}}%
    \put(0.0832078,0.13584291){\color[rgb]{0,0,0}\makebox(0,0)[lt]{\lineheight{1.25}\smash{\begin{tabular}[t]{l}RL Agent\end{tabular}}}}%
    \put(0.39573557,0.24512999){\color[rgb]{0,0,0}\makebox(0,0)[lt]{\lineheight{1.25}\smash{\begin{tabular}[t]{l}    Task 1 \\\end{tabular}}}}%
    \put(0.3982925,0.24413762){\color[rgb]{0,0,0}\makebox(0,0)[lt]{\lineheight{1.25}\smash{\begin{tabular}[t]{l}\\Controller\end{tabular}}}}%
    \put(0.40174748,0.11454873){\color[rgb]{0,0,0}\makebox(0,0)[lt]{\lineheight{1.25}\smash{\begin{tabular}[t]{l}   Task 2 \\\end{tabular}}}}%
    \put(0.40166699,0.11285937){\color[rgb]{0,0,0}\makebox(0,0)[lt]{\lineheight{1.25}\smash{\begin{tabular}[t]{l}\\Controller\end{tabular}}}}%
    \put(0,0){\includegraphics[width=\unitlength,page=1]{controller_block_diagram.pdf}}%
    \put(0.64778969,0.32010467){\color[rgb]{0,0,0}\makebox(0,0)[lt]{\lineheight{1.25}\smash{\begin{tabular}[t]{l}    Jacobian\\\end{tabular}}}}%
    \put(0.65127062,0.31884725){\color[rgb]{0,0,0}\makebox(0,0)[lt]{\lineheight{1.25}\smash{\begin{tabular}[t]{l}\\Transformation\end{tabular}}}}%
    \put(0,0){\includegraphics[width=\unitlength,page=2]{controller_block_diagram.pdf}}%
    \put(0.41024908,0.35663258){\color[rgb]{0.01176471,0.00784314,0.01960784}\transparent{0.94851202}\makebox(0,0)[lt]{\lineheight{1.25}\smash{\begin{tabular}[t]{l}$q$\end{tabular}}}}%
    \put(0.64678605,0.35363869){\color[rgb]{0.01176471,0.00784314,0.01960784}\transparent{0.94851202}\makebox(0,0)[lt]{\lineheight{1.25}\smash{\begin{tabular}[t]{l}$q$\end{tabular}}}}%
    \put(0.47916923,0.35666323){\color[rgb]{0.01176471,0.00784314,0.01960784}\transparent{0.94851202}\makebox(0,0)[lt]{\lineheight{1.25}\smash{\begin{tabular}[t]{l}$\dot{q}$\end{tabular}}}}%
    \put(0.53568387,0.25365467){\color[rgb]{0.01176471,0.00784314,0.01960784}\transparent{0.94851202}\makebox(0,0)[lt]{\lineheight{1.25}\smash{\begin{tabular}[t]{l}$F_1$\end{tabular}}}}%
    \put(0.79419297,0.20455535){\color[rgb]{0.01176471,0.00784314,0.01960784}\transparent{0.94851202}\makebox(0,0)[lt]{\lineheight{1.25}\smash{\begin{tabular}[t]{l}$\tau$\end{tabular}}}}%
    \put(0.53487109,0.1051615){\color[rgb]{0.01176471,0.00784314,0.01960784}\transparent{0.94851202}\makebox(0,0)[lt]{\lineheight{1.25}\smash{\begin{tabular}[t]{l}$F_2$\end{tabular}}}}%
    \put(0.20550979,0.10094945){\color[rgb]{0.01176471,0.00784314,0.01960784}\transparent{0.94851202}\makebox(0,0)[lt]{\lineheight{1.25}\smash{\begin{tabular}[t]{l}$q_{5,ref}, q_{7,ref}$\end{tabular}}}}%
    \put(0.27153767,0.26612112){\color[rgb]{0.01176471,0.00784314,0.01960784}\transparent{0.94851202}\makebox(0,0)[lt]{\lineheight{1.25}\smash{\begin{tabular}[t]{l}$x_{1,ref}$\end{tabular}}}}%
    \put(0.16535333,0.24137428){\color[rgb]{0.01176471,0.00784314,0.01960784}\transparent{0.94851202}\makebox(0,0)[lt]{\lineheight{1.25}\smash{\begin{tabular}[t]{l}$x_1, x_2, v$\end{tabular}}}}%
    \put(0.29790386,0.31905383){\color[rgb]{0.01176471,0.00784314,0.01960784}\transparent{0.94851202}\makebox(0,0)[lt]{\lineheight{1.25}\smash{\begin{tabular}[t]{l}$N$\end{tabular}}}}%
    \put(0.55259934,0.31926185){\color[rgb]{0.01176471,0.00784314,0.01960784}\transparent{0.94851202}\makebox(0,0)[lt]{\lineheight{1.25}\smash{\begin{tabular}[t]{l}$N$\end{tabular}}}}%
    \put(0.64412079,0.24886667){\color[rgb]{0.01176471,0.00784314,0.01960784}\transparent{0.94851202}\makebox(0,0)[lt]{\lineheight{1.25}\smash{\begin{tabular}[t]{l}$\bar{J}$\end{tabular}}}}%
    \put(0.089709,0.35449526){\color[rgb]{0.01176471,0.00784314,0.01960784}\transparent{0.94851202}\makebox(0,0)[lt]{\lineheight{1.25}\smash{\begin{tabular}[t]{l}$q$\end{tabular}}}}%
    \put(0.15068367,0.35408725){\color[rgb]{0.01176471,0.00784314,0.01960784}\transparent{0.94851202}\makebox(0,0)[lt]{\lineheight{1.25}\smash{\begin{tabular}[t]{l}$\dot{q}$\end{tabular}}}}%
    \put(0.08687947,0.02877967){\color[rgb]{0.01176471,0.00784314,0.01960784}\transparent{0.94851202}\makebox(0,0)[lt]{\lineheight{1.25}\smash{\begin{tabular}[t]{l}$q$\end{tabular}}}}%
    \put(0.13664659,0.02868909){\color[rgb]{0.01176471,0.00784314,0.01960784}\transparent{0.94851202}\makebox(0,0)[lt]{\lineheight{1.25}\smash{\begin{tabular}[t]{l}$\dot{q}$\end{tabular}}}}%
    \put(0,0){\includegraphics[width=\unitlength,page=3]{controller_block_diagram.pdf}}%
  \end{picture}%
\endgroup%

%% file: sections/4-rl.tex
\label{sec:rl}
In this section, the hierarchical control framework is integrated with a reinforcement learning (RL) agent to achieve desired reference trajectories for the suspended aerial manipulation system. The Soft Actor-Critic (SAC) algorithm (\cite{haarnoja2018soft}) is selected for its ability to handle continuous action spaces, which are inherent to the control of robotic manipulators. Furthermore, SAC balances exploration and exploitation, ensuring effective learning and robust policy performance.

Actor-critic methods combine value-based and policy-based reinforcement learning principles, addressing some of the limitations of standalone approaches. As an Actor-Critic method, SAC utilizes a policy gradient to optimize a parameterized policy. Unlike REINFORCE, see \cite{arulkumaran2017deep}, which updates policies based on episodic returns, SAC employs a Critic to estimate the action-value function, enhancing sample efficiency. The policy gradient update in SAC incorporates the advantage function; see \cite{haarnoja2018soft}:
\begin{equation}
     \nabla_{\vec{\theta}} J(\pi_{\vec{\theta}}) = \mathbb{E}\Big[\nabla_{\vec{\theta}} \log \pi_{\vec{\theta}} (\mathbf{u}_k| \mathbf{x}_k) A^\pi(\mathbf{x}_k,\mathbf{u}_k)\Big],
     \label{eq:advantage_actor_critic}
\end{equation}
where $A^\pi(\mathbf{x}_k,\mathbf{u}_k) = Q^\pi(\mathbf{x}_k,\mathbf{u}_k) - V^\pi(\mathbf{x}_k)$ represents the advantage function, with $\mathbf{x}_k$ and $\mathbf{u}_k$ denoting the observation and the action space vector at the $i^{th}$ iteration, respectively. This formulation ensures stable training dynamics and improved performance in continuous action spaces.

To train the RL agent, we consider the observation space as the joints $\mathbf{q}$ and its velocities $\mathbf{\dot{q}}$. The action space consists of the two elbow joints $q_5$ and $q_7$. We choose the reward function as the exponential error of the platform joint $q_2$ along the swing-up axis. As mentioned previously, the primary task $\mathbf{x}_1$ does not contribute to the swing-up that is accomplished using the secondary task $\mathbf{x}_2$. Additionally, due to the involvement of certain joints $q_4$ and $q_6$ perpendicular to the swing-up axis, a zero reference is commanded to them. In this work, we demonstrate the swing-up motion by using minimal thrust force from the platform. Therefore, the task force $\mathbf{F}_2$ corresponding to the joint $q_2$ is used to counteract its corresponding gravity compensation $g_2$ in \eqref{eq:joint_torque}. For similar reasons, the component $\tau_{u,2}$ of the cross-coupling compensation torque $\bm{\tau}_u$ corresponding to $q_2$ is also considered as zero. The elbow joint $q_9$ is not considered in the second task due to its comparatively low inertia and possible conflict with the primary task. 

The proposed reinforcement learning framework thus utilizes an observation space 
\(\begin{bmatrix} \mathbf{q} & \dot{\mathbf{q}} \end{bmatrix}^T\) and an action space 
\(\begin{bmatrix} q_{5,\text{ref}} & q_{7,\text{ref}} \end{bmatrix}^T\), where 
\(q_{5,\text{ref}}\) and \(q_{7,\text{ref}}\) represent the reference commands for 
the corresponding joints. The agent is rewarded according to the exponential function 
\( e^{-5 \lvert q_2 - 1 \rvert} \), thus favoring actions that bring \( q_2 \) close 
to \SI{1}{\radian}. To ensure proper task execution, the references 
\(\mathbf{x}_{1,\text{ref}} = \begin{bmatrix}0 & 0 & 0 & -\tfrac{\pi}{2} & 0\end{bmatrix}^T\) 
and 
\(\mathbf{x}_{2,\text{ref}}[2{:}5] = \begin{bmatrix}0 & q_{5,\text{ref}} & 0 & q_{7,\text{ref}}\end{bmatrix}^T\), where [2:5] indicates elements 2 through 5 of $x_{2,ref}$
are defined for the respective tasks, with the thrust force of the platform commanded
to cancel out the gravity compensation \( g_2 \) and the cross-coupling Coriolis compensation $\tau_{u,2}$ through the task force 
\(\mathbf{F}_{2}[1]\) corresponding to $q_2$. This setup ensures stable platform behavior and effective 
regulation of the target joints through the learned policy.

%% file: sections/5-experiments.tex
\label{sec:exp}
This section presents the numerical simulation results and analyzes the swing-up maneuver with our proposed approach. We considered the mass of the suspended aerial platform to be 4 kg, with its principal moment of inertia along its $x$, $y$, and $z$ axis being 0.0646, 0.0646 and 0.0682 \SI{}{\kilogram\square\meter}, respectively, similar to our previous work in \cite{das2023}. The platform is attached using a rigid cable of length \SI{1}{\meter}, with its mass being \SI{0.3}{\kilogram}. The mass of the 7-DoF manipulator is considered as \SI{8.2}{\kilogram}, with the maximum reach of the end-effector being \SI{0.9}{\meter}. We performed our analysis in the Mujoco Simulation environment (\cite{todorov2012mujoco}), with a sampling time of \SI{1000}{\hertz}. We utilized the Stable-baseline3 RL library (\cite{raffin2021stable}), which is based on PyTorch, to train the RL agent. Both the actor and critic networks contain four hidden layers, each containing 256 neurons. We trained the networks for a total of around 1300 episodes, with each episode lasting \SI{15}{\second}. The training is accomplished in a custom-made environment based on OpenAI Gym. The total training time is roughly \SI{100}{\minute}, with 30 environments being used in parallel to speed up the training time.  \par 
\begin{figure}[h]
    \centering
\includegraphics[width = 0.45 \textwidth]{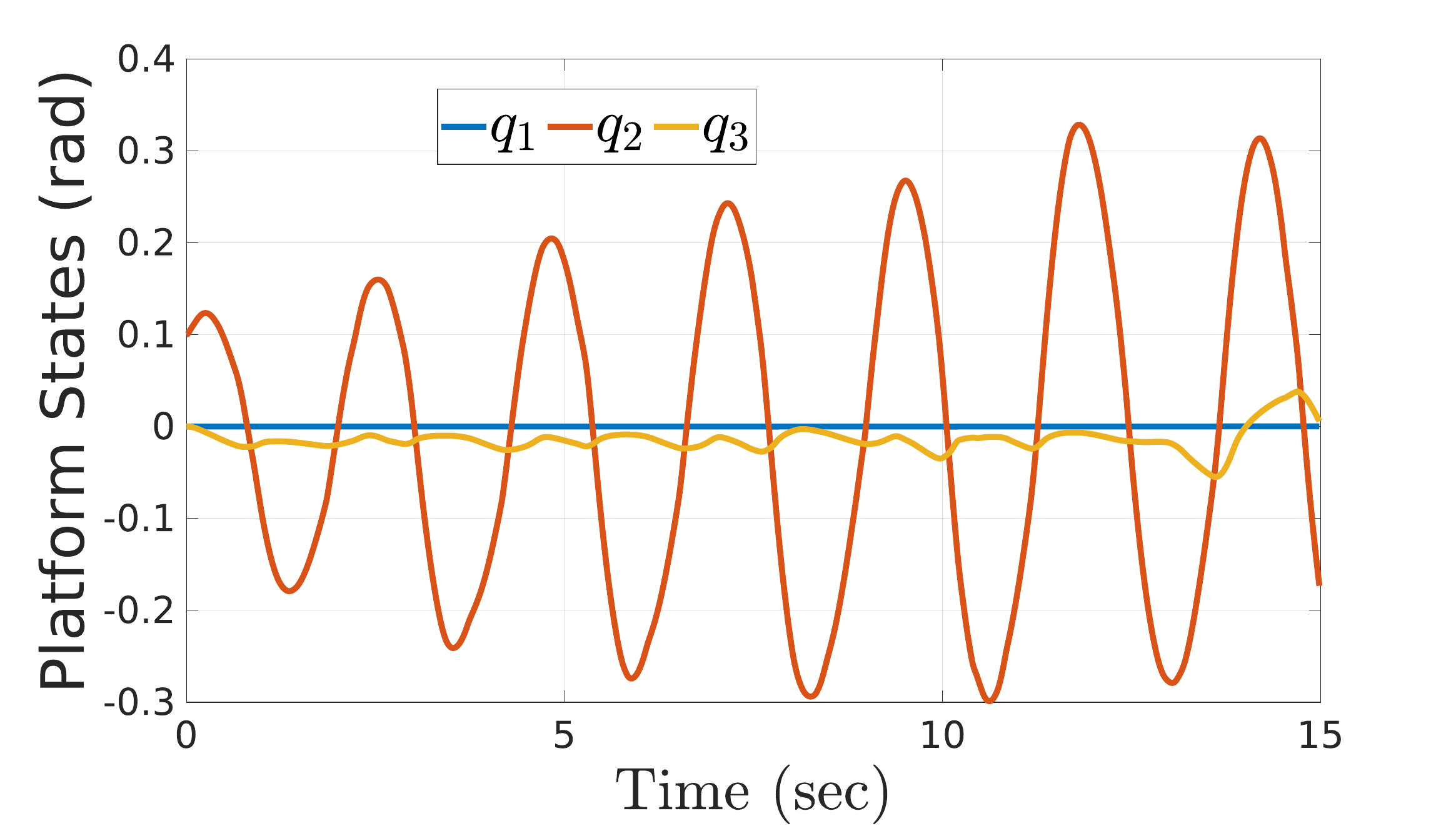}
    \caption{Swing-up behaviour of the platform.}
    \label{fig:plat_beh}
\end{figure} 
\begin{figure}[h]
    \centering
\includegraphics[width = 0.45 \textwidth]{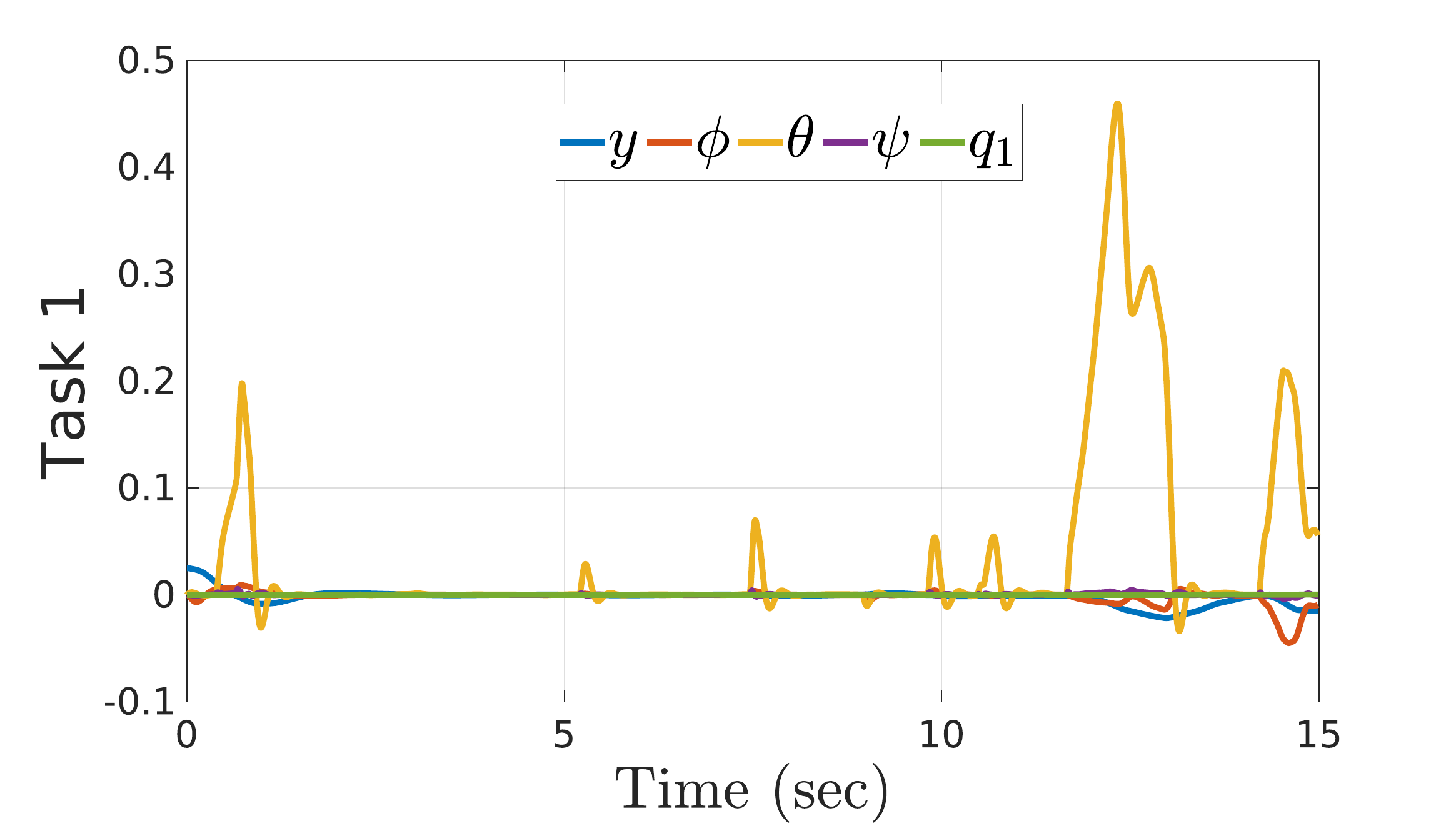}
    \caption{Behaviour of task $\bm{x}_1$, with $y$ expressed in \SI{}{\meter}, while $\phi$, $\theta$, $\psi$ and $q_1$ are expressed in \SI{}{\radian}.}
    \label{fig:task}
\end{figure} 

 \begin{figure}[h]
    \centering
\includegraphics[width = 0.45 \textwidth]{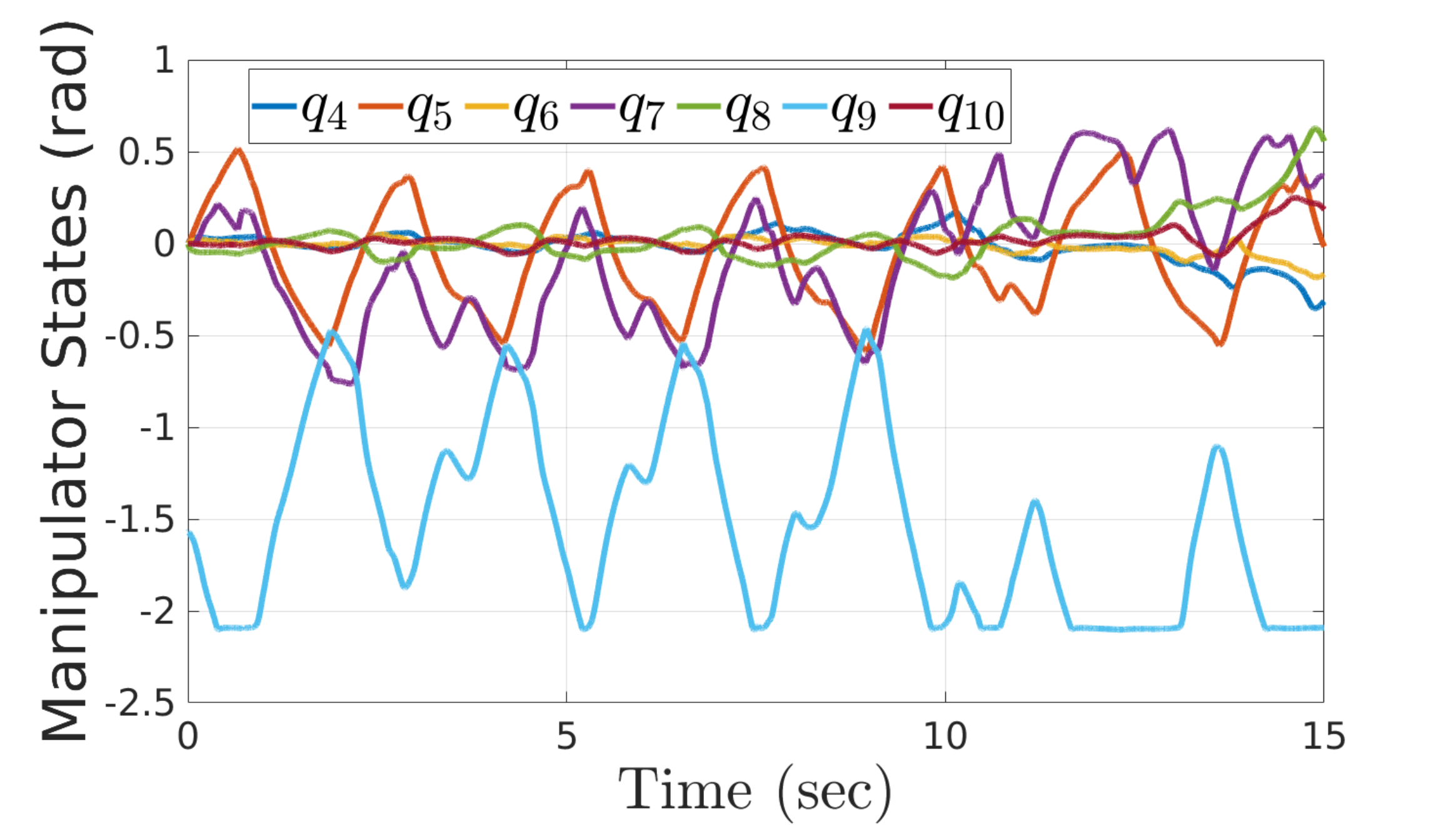}
    \caption{Behaviour of the manipulator.}
    \label{fig:manip}
\end{figure} 
\begin{figure*}[t]
    \centering
\includegraphics[width =0.91 \textwidth]{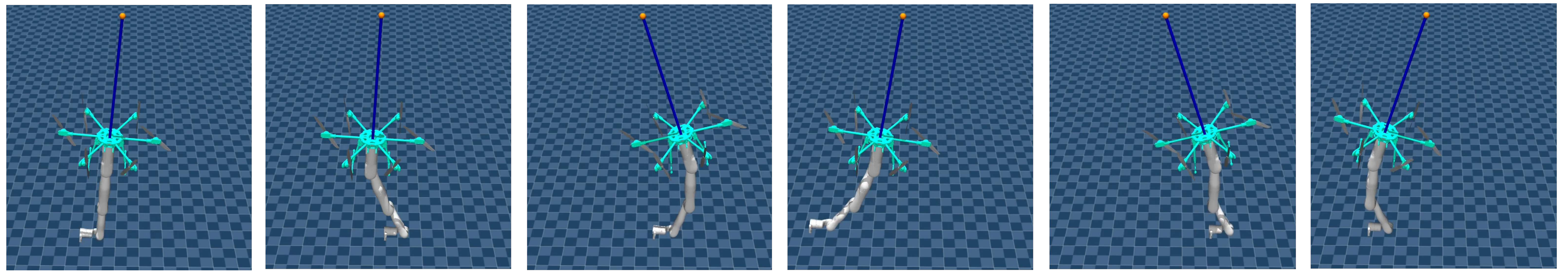}
    \caption{Sequence of the swing-up experiments at different times t = 0 s, 5.3 s, 8.1 s, 9.2 s, 12.9 s, and 14.2 s, displayed from left to right.}
    \label{fig:exp_sequence}
\end{figure*} 

 \begin{figure}[h]
    \centering
\includegraphics[width = 0.45 \textwidth]{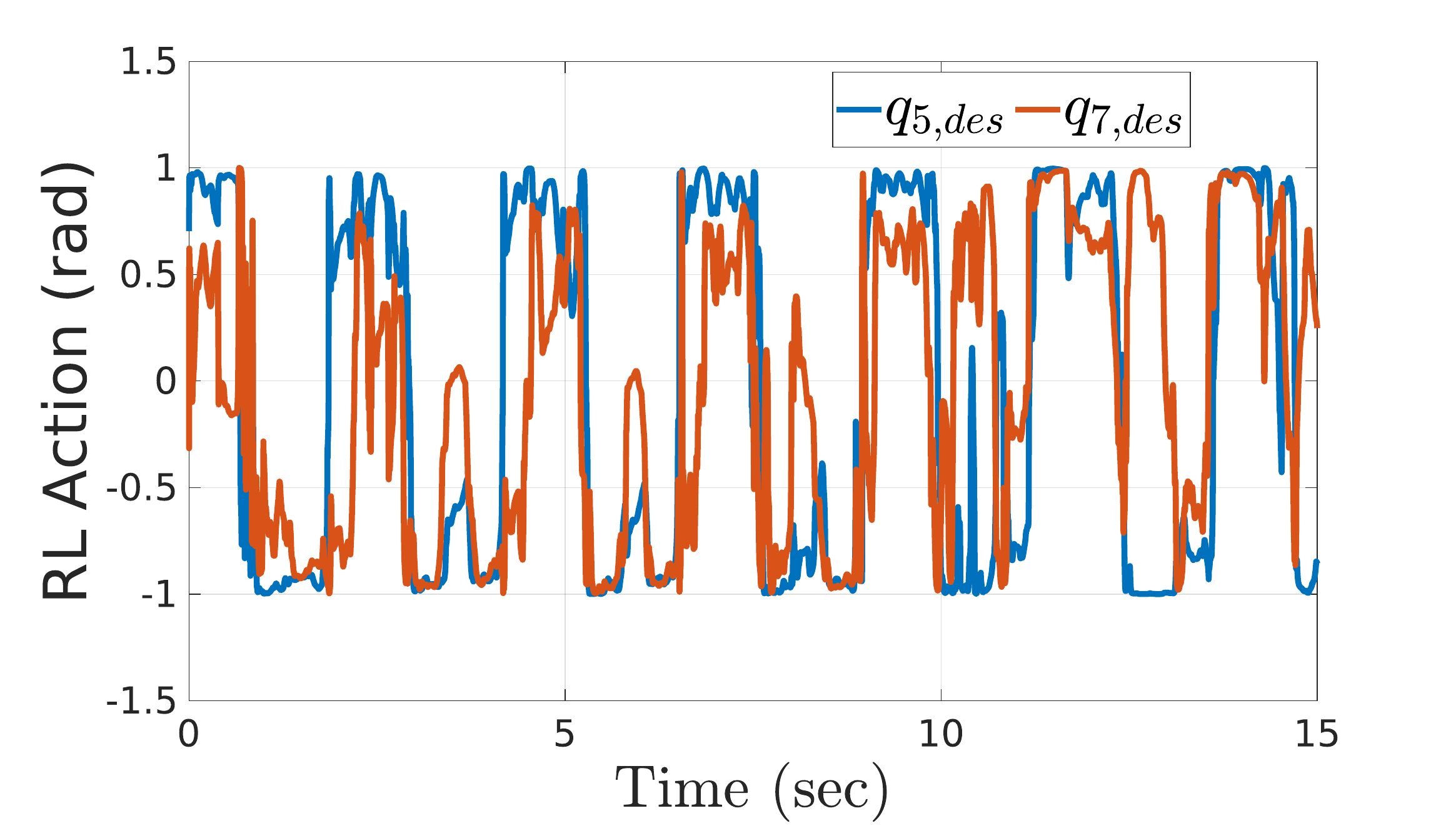}
    \caption{Commanded angles by the RL agent for the joints $q_5$ and $q_7$.}
    \label{fig:rl_command}
\end{figure} 
 
 \begin{figure}[h]
    \centering
\includegraphics[width = 0.45 \textwidth]{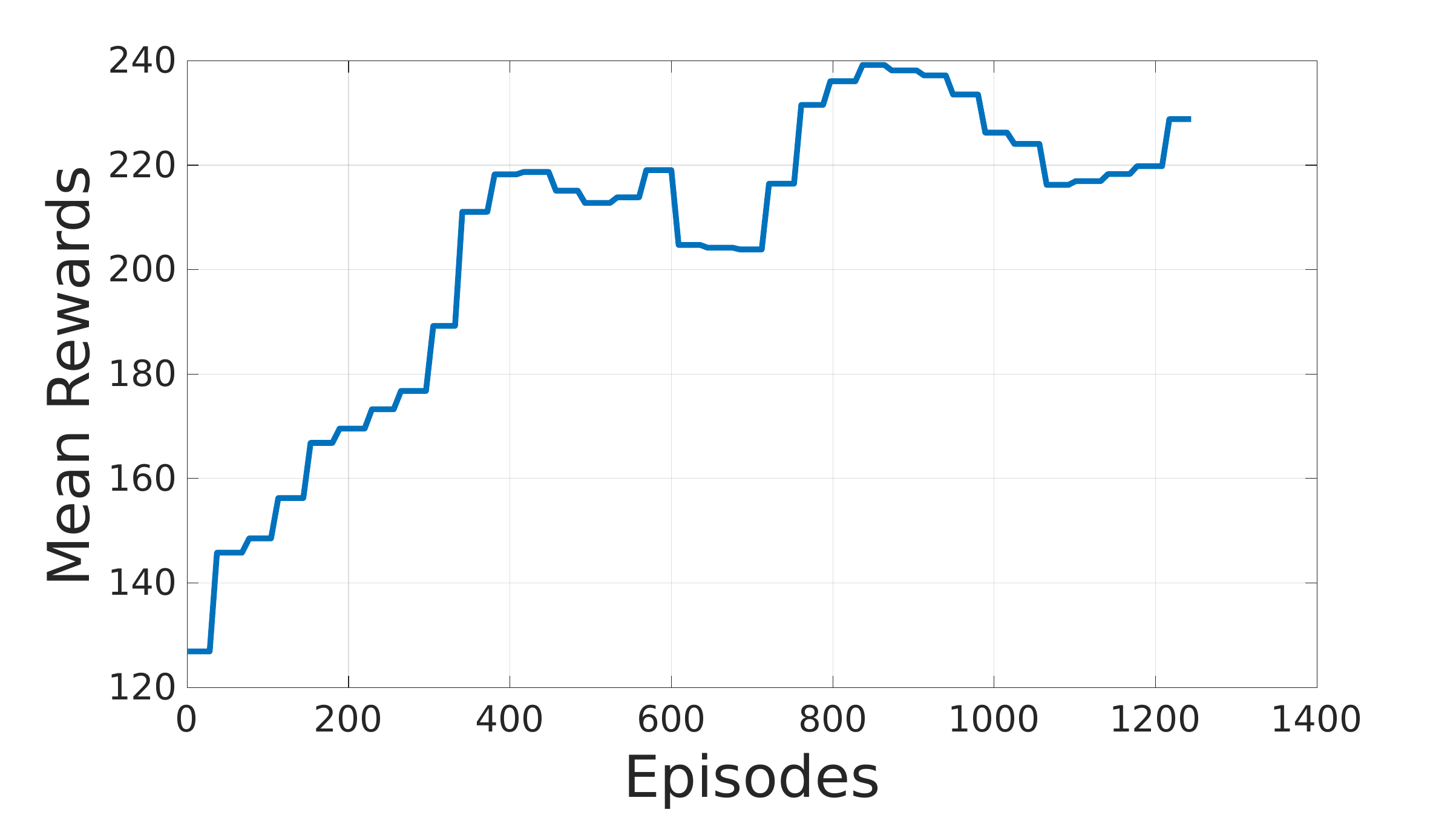}
    \caption{Mean episodic rewards during training of the RL agent.}
    \label{fig:reward_explore}
\end{figure} \par
\begin{figure}
\begin{subfigure}{.5\textwidth}
  \centering
  \includegraphics[width=0.9\linewidth]{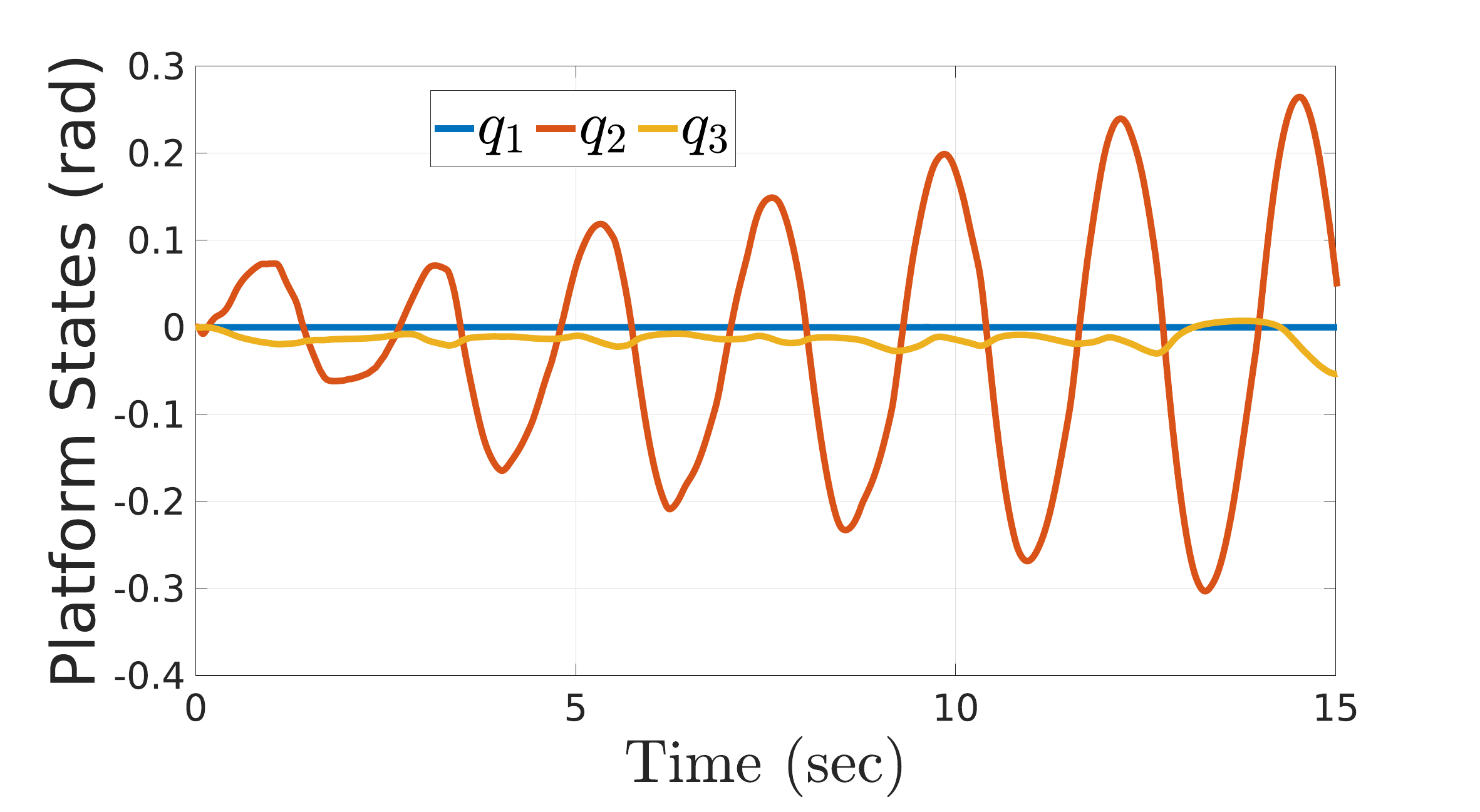}
  \caption{{Swing-up behaviour of the platform with $q_{2,0}$ = \SI{0}{\radian}.}}
  \label{fig:sub-robust-wind1}
\end{subfigure}
\begin{subfigure}{.5\textwidth}
  \centering
\includegraphics[width=0.9\linewidth]{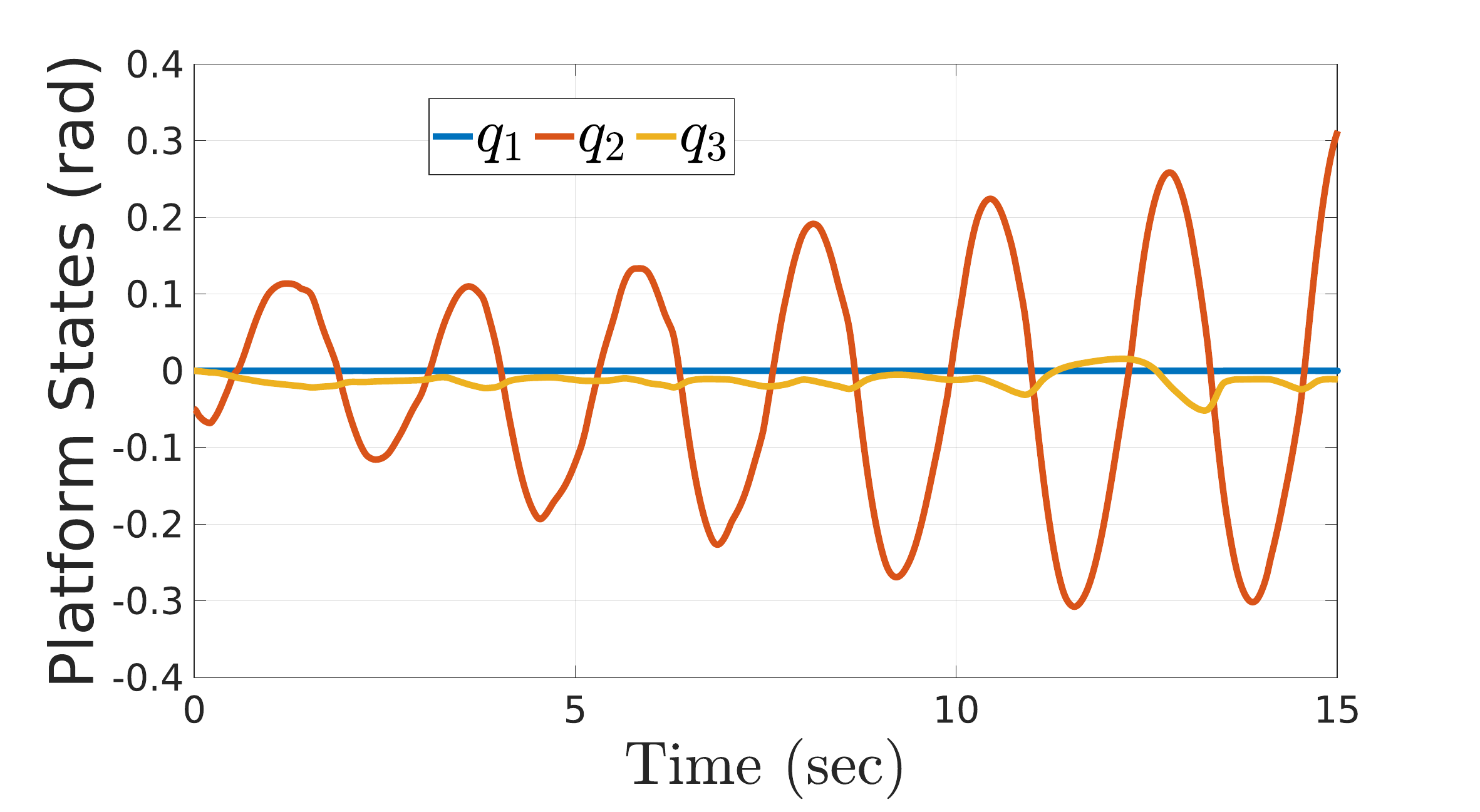}
  \caption{{Swing-up behaviour of the platform with $q_{2,0}$ = \SI{-0.05}{\radian}.}}
  \label{fig:sub-robust-wind2}
\end{subfigure}
\caption{{Swing-up behaviour of the platform with different initial conditions.}}
\label{fig:differnt_initi}
\end{figure}

 
In this work, we command the task $\bm{x}_2$ to execute the swing-up maneuver in the nullspace of the primary task $\bm{x}_1$. Additionally, to aid in the swing-up maneuver, we considered the initial angle $q_{2,0}$ for the joint $q_2$ during both the training and evaluation phases as \SI{0.1}{\radian}, which can be achieved by the propulsion force of the platform. The sequence of the swing-up experiment is shown in Fig. \ref{fig:exp_sequence}.
From Fig. \ref{fig:plat_beh}, we observe that the platform can reach the swing-up angle of up to \SI{0.32}{\radian} in a time-period of \SI{15}{\second}.  The joint angles commanded 
 by the RL agent is limited between \SI{-1}{\radian} and \SI{1}{\radian} to avoid any aggressive swing-up maneuver (Fig. \ref{fig:rl_command}). The RL agent commands action to the secondary task in the nullspace of the primary task, and therefore do not influence it due to the involved dynamical decoupling.  The mean episodic rewards during the training of the RL agent are shown in Fig. \ref{fig:reward_explore}. The behavior of the task $\mathbf{x}_1$  is plotted in Fig. \ref{fig:task}. We observe that the tracking error is not zero throughout the swing-up maneuver, which can possibly be reduced by increasing the actuator bandwidth and it saturation limits. The behavior of the manipulator is plotted in figure \ref{fig:manip}, where we observe the motion of the joints $q_5$ and $q_7$ utilized for the swing-up maneuver. We also observe that the joint $q_9$ is primarily used for maintaining the end-effector orientation along the $y$-$z$ plane. Furthermore, in order to demonstrate the robustness of our trained network, we also demonstrate the swing-up maneuver of the platform from different initial configurations, as shown in Fig. \ref{fig:differnt_initi}.

%% file: sections/6-conclusion.tex
\label{sec:con}
In this work, we integrated a reinforcement learning agent to adjust the secondary task references in the null space of the higher-priority task. Our framework is based on a hierarchical control framework, which ensures that the primary tasks related to both the orientation and the $y$ axis position of the end-effector, and also the platform's yaw angle, are not interfered with by the execution of the secondary tasks related to the swing-up maneuver. We utilized an off-policy RL agent based on the state-of-the-art SAC algorithm to obtain the reference setpoints of the two elbow joints of the manipulator and we successfully demonstrated the swing-up maneuver of the suspended aerial manipulation system.  \par 
We will implement the RL agent on the physical system as future work. This would further necessitate domain randomization to ensure the robustness of the learned policy to unknown system dynamics and sensor noise.
We would also like to analyze the use of the thrusters of the platform to further aid the swing-up maneuver by integrating it into the RL agent.